\documentclass{article}

\usepackage[preprint,nonatbib]{neurips_2025}
\usepackage[utf8]{inputenc} % allow utf-8 input
\usepackage[T1]{fontenc}    % use 8-bit T1 fonts
\usepackage{hyperref}       % hyperlinks
\usepackage{url}            % simple URL typesetting
\usepackage{booktabs}       % professional-quality tables
\usepackage{amsfonts}       % blackboard math symbols
\usepackage{nicefrac}       % compact symbols for 1/2, etc.
\usepackage{microtype}      % microtypography
\usepackage{graphicx} % Required for including images
\usepackage[ruled,vlined]{algorithm2e} % for the algorithms
\usepackage{amsmath} % Include the amsmath package
\usepackage{mathdots} % for diagonal dots in matrix
\usepackage[title]{appendix}
\usepackage[colorinlistoftodos]{todonotes}
\usepackage{todonotes} % for adding to do notes
\usepackage{multirow}
\usepackage{booktabs}
\usepackage{xcolor}
\usepackage{colortbl}
\usepackage{rotating}
\usepackage{enumitem}

\title{When Reasoning Beats Scale: A 1.5B Reasoning Model Outranks 13B LLMs as Discriminator }

\author{%
  Md Fahim~Anjum\thanks{Alternate email: dr.fahim.anjum@gmail.com} \\
  Department of Neurology\\
  University of California San Francisco\\
  San Francisco, CA 94143 \\
  \texttt{fahim.anjum@ucsf.edu} \\
}

\begin{document}

\maketitle

\begin{abstract}
Large Language Models (LLM) with reasoning capabilities offer a promising path for improving candidate evaluation in planning frameworks, but their relative performance against traditional non‑reasoning models remains largely underexplored. In this study, we benchmark a distilled 1.5B parameter reasoning model (DeepSeek‑R1) against several state‑of‑the‑art non‑reasoning LLMs within a generator-discriminator LLM planning framework for the text-to-SQL task. 
For this, we introduce a novel method for extracting soft scores from the chain-of-thought (CoT) outputs from reasoning that enables fine-grained ranking of candidates. Our central hypothesis is that reasoning models are more effective discriminators than non-reasoning LLMs. Our results show that distilled DeepSeek-R1-1.5B achieves up to \textbf{87\%} higher F1 and \textbf{3.7\%} better discrimination accuracy than CodeLlama-7B, as well as \textbf{3.7\%} higher execution accuracy than CodeLlama-13B, despite having significantly fewer parameters. Furthermore, we find that there is a limit to the logical capabilities of reasoning models, and only providing more context or allowing more compute budget for reasoning is not enough to improve their discrimination performance. Finally, we demonstrate that, unlike non-reasoning LLMs, reasoning models find generation more challenging than discrimination and may underperform as generators compared to smaller non-reasoning LLMs. Our work highlights the potential of reasoning models as discriminators in agentic frameworks, far outweighing their capabilities as generators, offering insights into their optimal role within LLM planning infrastructures.
\end{abstract}

\section{Introduction}

Planning is a fundamental part of intelligent decision-making. Significant research efforts have focused on developing approaches to enable effective and human-like planning, including multi-phase and multi-agent frameworks~\cite{plangen, xie2024, newellsimon1956planning, russel2010}. The recent integration of large language models (LLMs) into such systems has spurred significant interest in LLM-based planning frameworks to enhance the multi-step problem-solving abilities using structured planning, which have shown strong performance across a variety of tasks~\cite{li-etal-2023-making, madaan2023selfrefine, yao2023tot, yao2023react, zhou2023language, feng2024alphazerolike, plangen, xie2024}. Despite differences in implementation and design, most of these frameworks follow a three-stage process: (1) searching for potential action sequences, (2) predicting expected outcomes, and (3) selecting the sequence that maximizes the expected utility~\cite{russel2010, planning_brain}. Specifically, LLMs first generate candidate actions (\textit{generation}), evaluate their consequences using themselves, internal simulations, or another model (\textit{discrimination}), and finally integrate these outcomes for optimal solutions (\textit{planning}). This structure aligns closely with how LLMs approach multi-step tasks such as mathematical reasoning~\cite{wei2022cot}, multi-hop question answering~\cite{yao2023react}, and code generation~\cite{yang2023intercode}. Performance of conventional non-reasoning LLMs in such frameworks has been studied comprehensively \cite{main}.
However, recent developments on LLMs designed for inference-time reasoning via internal deliberation show promising performance across various tasks. Augmented with chain-of-thought (CoT) reasoning \cite{wei2023chainofthoughtpromptingelicitsreasoning}, these models offer a structured approach to complex tasks. Large reasoning models such as OpenAI o-series \cite{openai2024openaio1card}, Gemini 2.0 Flash Thinking \cite{google2024gemini}, DeepSeek-R1 \cite{deepseekai2025deepseekr1incentivizingreasoningcapability}, LIMO \cite{ye2025limoreasoning} , and Qwen QwQ-32B \cite{qwen2025qwq32b} have shown impressive stepwise reasoning and state-of-the-art (SOTA) performance across complex benchmarks, far surpassing non-reasoning LLMs  \cite{wu2025phdknowledgerequiredreasoning}.
Notably, DeepSeek-R1 \cite{deepseekai2025deepseekr1incentivizingreasoningcapability} has achieved SOTA performance on par with OpenAI’s models despite low training cost. With logical and accurate outputs,  it consistently outperforms Qwen and LLaMA series models, even under quantization and distillation \cite{zhao2025quantifyingcapabilityboundarydeepseek, wu2025agenticreasoningreasoningllms}. 
Despite the success of reasoning models over traditional non-reasoning LLMs, only a handful of studies have explored their integration into planning or multi-agent frameworks \cite{wu2025agenticreasoningreasoningllms, saha2025learningplanreason, chen2025reasoningerasurveylong}. To the best of our knowledge, there has been no systematic investigation into how reasoning models compare to conventional non-reasoning LLMs within the generator-discriminator LLM planning architecture.

In this work, we examine the role of a reasoning model within a generator-discriminator LLM planning agentic framework applied to the text-to-SQL generation task. For this, we use a distilled 1.5B model of DeepSeek-R1 and hypothesize that reasoning models may serve as more effective discriminators and conduct a systematic investigation guided by the following research questions: 

\noindent \textbf{(1)} \textit{How does incorporating a reasoning LLM as the discriminator influence the accuracy and overall performance of the agentic framework? }

\noindent \textbf{(2)} \textit{What is the impact of inference-time CoT compute budget and contextual augmentation on the discrimination performance? }

\noindent \textbf{(2)} \textit{ How does a reasoning model perform in the generator role, and is it better at generation than discrimination?}

To this end, we analyze and comprehensively evaluate DeepSeek-R1's discrimination as well as generation abilities within our LLM planning framework and compare the performance with several SOTA non-reasoning LLMs. 
An additional challenge arises in the extraction of soft scores from CoT reasoning outputs. Traditional non-reasoning LLMs can easily provide logit-based scores for discrimination, but in reasoning models, the arbitrarily long and unstructured nature of the CoT trace complicates soft score extraction. To address this, we propose a novel approach for deriving soft scores from the reasoning outputs, enabling more refined discrimination.

Our work evaluates LLMs' ability to discriminate correct and incorrect SQL queries by re-labeling oracle-generated outputs, measuring both classification and ranking. Discriminator performance is tested in a naive setting and an enhanced setting that filters by executability via environment-based validation.  We further test whether LLMs can select correct queries from dynamically generated candidates. We analyze DeepSeek-R1's reasoning quality through lexical, structural, and semantic diversity and examine three key factors influencing its discrimination performance: (1) increased test-time compute budget, (2) prompt augmentation with database schema context, and (3) the use of binary decisions vs. logit-based soft scores. Finally, we evaluate LLMs' performance as generators in the full LLM planning pipeline.
Our contributions in this work are as follows:

\noindent \textbf{(1)} We provide a systematic comparison of reasoning and non-reasoning LLMs within a generator-discriminator LLM planning framework for the text-to-SQL task.  

\noindent \textbf{(2)} We introduce a novel use of reasoning models as discriminators by deriving soft scores from their outputs for more fine-grained candidate ranking. 

\noindent \textbf{(3)} We show that DeepSeek-R1 excels at discrimination, outperforming larger non-reasoning models. In particular, we find that a 1.5B distilled DeepSeek-R1 model achieves  $87\%$ higher F1 as well as $3.7\%$ better discrimination accuracy than CodeLlama-7B and $3.7\%$ higher execution accuracy than CodeLlama-13B. Furthermore, logit-based soft scoring vs. binary true/false discrimination yields minimal differences ($<1.5\%$). 

\noindent \textbf{(4)} We find that reasoning models have limited logical capacity and adding more context or compute budget yields diminishing returns (e.g., <0.4\% gain beyond 1024 tokens), while extremely low budgets severely degrade performance ($<2\%$ accuracy, $>94\%$ failure).

\noindent \textbf{(5)} Finally, we show that while DeepSeek-R1 is a strong discriminator, it underperforms as a generator, even compared to smaller non-reasoning LLMs. We find that, similar to humans, generation is more challenging than discrimination for reasoning models \cite{gu-etal-2023-dont} , in contrast to non-reasoning LLMs~\cite{main, west2024paradox, huang2024llm_selfcorrect, wang-etal-2023-chatgpt-defend}.

In summary, our work investigates the role of DeepSeek-R1 within an LLM planning framework for text-to-SQL task. 
Through this study, we seek to advance our understanding of reasoning-based LLMs and inform future efforts in designing robust, efficient, and scalable autonomous systems.

\section{Related Work}\label{sec:priorworks}
In a generator-discriminator LLM planning framework, discrimination is often seen as easier than generation for human and AI agents \cite{gu-etal-2023-dont}. However,  West et al.  hypothesized that generative models such as LLMs may lack discrimination abilities to match their generation capabilities \cite{west2024paradox}.
This aligns with findings from other studies, which suggest that non-reasoning LLMs may misidentify correct self-generated plans as incorrect without external feedback \cite{huang2024llm_selfcorrect, wang-etal-2023-chatgpt-defend}. 
Huang et al. observed that self-correction performance gains may depend on high-quality external feedback  \cite{huang2024llm_selfcorrect} 
Chen et al. focused on non-reasoning LLMs' discrimination abilities and their impact on several planning methods and found a strong correlation between discrimination accuracy and overall task performance \cite{main}.

In the context of test-time compute for reasoning models,  Zeng et al. observed an inverse scaling phenomenon where accuracy declines as CoT length increases \cite{time-scale-o1}. They also found that correct solutions are consistently shorter than incorrect ones, suggesting longer CoTs may actually harm performance.  Other studies have shown that exceeding a model's inherent reasoning capacity with longer CoTs leads to error accumulation as well as performance degradation  \cite{jin-etal-2024-impact, chen2024unlocking, wu2025lessunderstandingchainofthoughtlength}
and simply increasing inference-time compute budget does not guarantee performance improvement \cite{parashar2025inferencetimecomputationsllmreasoning, wang2025examiningfalsepositivesinference}.
Moreover, recent studies have found that optimal reasoning length depends on the model's capability as well as on the complexity of the task. \cite{wu2025phdknowledgerequiredreasoning, setlur2025scalingtesttimecomputeverification, balachandran2025inferencetimescalingcomplextasks}

Unlike these prior studies that focus either on non-reasoning models for planning or reasoning models' CoT without any agentic framework, our work investigates the role of a reasoning model within the LLM planning framework and examines the impact of inference-time compute budget on performance. 

\section{Experimental Setup}\label{sec:exp}
\subsection{Task and Dataset}
We evaluate the text-to-SQL task using the Spider dataset \cite{yu-etal-2018-spider}. Following prior work \cite{main}, we use a resource-efficient subset of 400 examples from the Spider development set, uniformly sampled across difficulty levels (100 each from easy, medium, hard, and extra-hard categories). The full Spider development set consists of 1,034 examples, with 24\% easy (248), 43.1\% medium (446), and 32.9\% hard or extra hard (340). The random sampling procedure of the examples is described in \cite{main}.

\subsection{Language Models}
We use \textit{DeepSeek-R1-Distill-Qwen-1.5B} \cite{distilledr1} as the primary reasoning model, referred to as `Distill-R1' throughout this work. To compare against non-reasoning architectures, we select a diverse set of open-source LLMs spanning various sizes: \textit{TinyLlama-1.1B-Chat-v1.0} \cite{tinyllama}, \textit{DeepSeek-Coder-1.3B-Instruct} \cite{deepseek_coder_1.3b_instruct}, \textit{Stable-Code-3B} \cite{stable-code-3b}, \textit{CodeLlama-7B-Instruct}, and \textit{CodeLlama-13B-Instruct} \cite{codellama}. For brevity, we omit the `Instruct' suffix in model names.

\subsection{Framework for LLM Planning}
\begin{figure}[htbp!]
	\centering
	\includegraphics[width=13.5cm,height=10.5cm,clip,keepaspectratio]{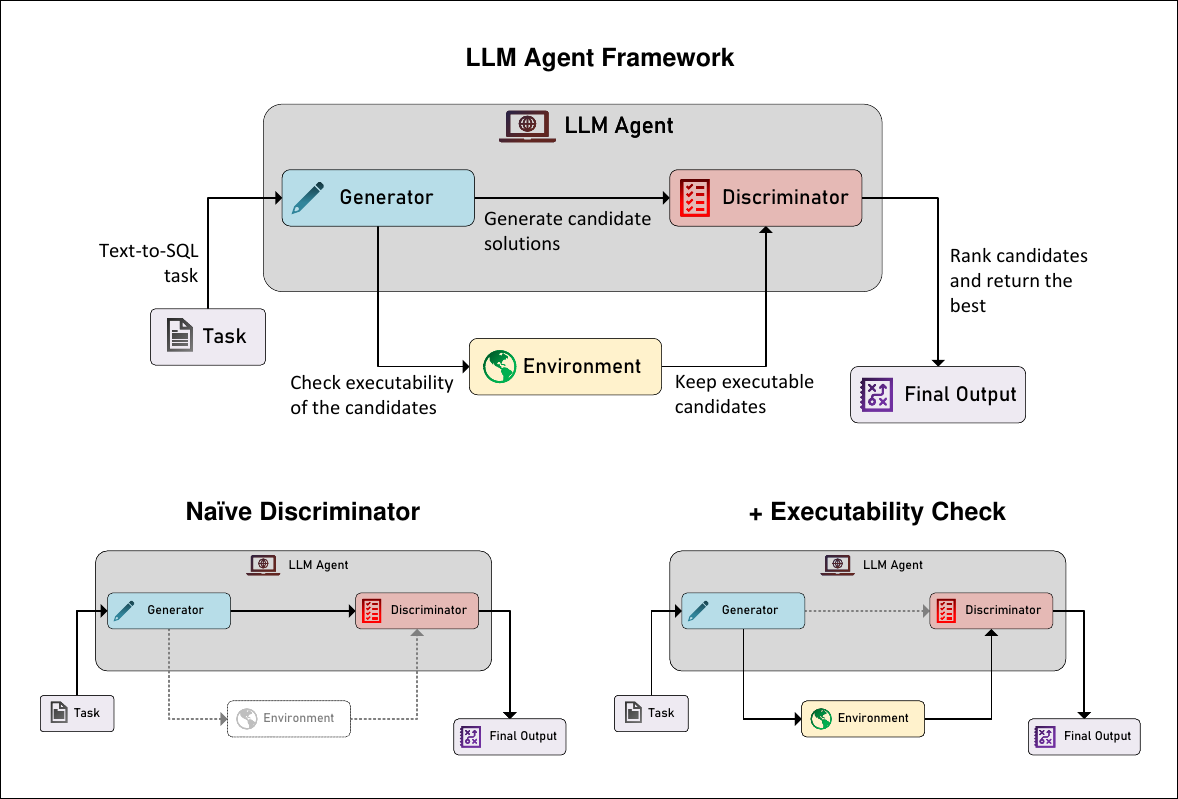}
	\caption{Framework of LLM planning for Text-to-SQL task utilized for this study. For further information, see \cite{main}. }
	\label{fig:overview}
\end{figure}

As illustrated in Figure~\ref{fig:overview}, we adopt a generator-discriminator framework \cite{main}, where the generator LLM proposes candidate solutions or actions and the discriminator LLM evaluates them. The planning module ranks candidates based on the evaluation outcomes and orchestrates the interaction between the two models. In this section, we detail each of these components and their specific implementation for text-to-SQL parsing.

\subsubsection{Generator}

We use the generator LLM to produce candidate SQL queries for user-provided natural language questions via zero-shot prompting. Database tables are included as context, and a fixed prompt template is employed to guide candidate generation. For non-reasoning models, we prompt with the instruction \texttt{`Given database schema and a question in natural language, generate the corresponding SQL query.''}. For the reasoning model (Distill-R1), the instruction is \texttt{``Answer the question by generating a valid sqlite SQL query. Final answer must start with SELECT.''} which allows the model to use reasoning for generating the query. The prompt template in this case is provided in markdown format, facilitating the extraction of the final SQL query generated by the model. Example prompts for both reasoning and non-reasoning models are given in the Appendix (Table \ref{tab:example3} and Table \ref{tab:example4}). 

We generate text-to-SQL outputs using the HuggingFace implementation \cite{wolf-etal-2020-transformers}, leveraging its model checkpoint and generation function. The hyperparameters are set as follows: $\mathtt{max\_length} = 300$, $\mathtt{temperature} = 0.2$, and $\mathtt{num\_return\_sequences} = 5$. When a single output is generated without planning or discriminator feedback ($\mathtt{num\_return\_sequences} = 1$), we refer to it as \textit{greedy generation}.

\subsubsection{Discriminator}

Given some candidate solutions, we formulate the discrimination task as a binary question answering \cite{kadavath2022language, ke-etal-2023-decompeval} , and generate a discrimination score for each candidate.

\noindent \textbf{Non-reasoning Models:} 
For conventional LLMs, we follow a straightforward scoring approach: the input prompt is passed to the model, and the discrimination score is computed as the token-level probability of generating \texttt{``Yes''} as the next token by applying a softmax function over the final logits \cite{main}. Specifically, we prompt the LLMs with the question \texttt{``Is the SQL program correct given the utterance?''} and extract the softmax probability assigned to \texttt{``Yes''} as the output score. An illustrative example is provided in the Appendix (Table~\ref{tab:example2}).

\begin{figure}[htbp!]
	\centering
	\includegraphics[width=13.5cm,height=10.5cm,clip,keepaspectratio]{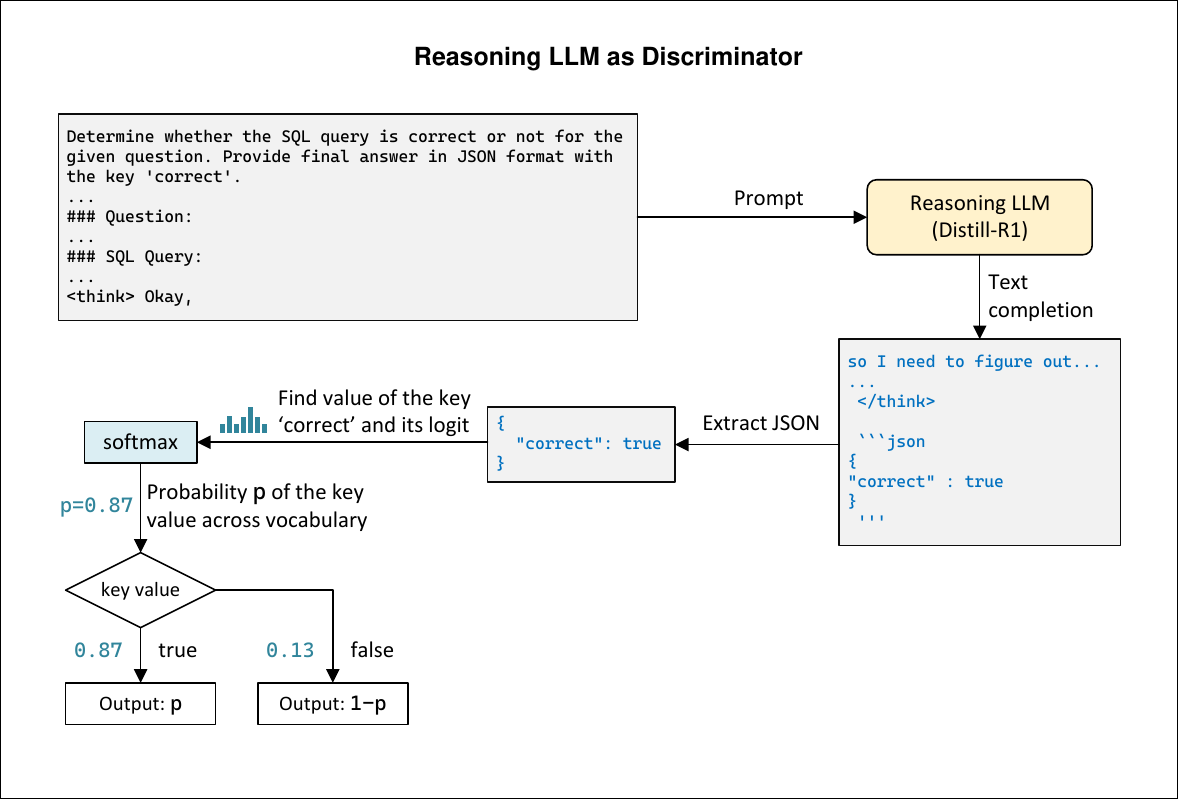}
	\caption{Flowchart illustrating our approach for obtaining soft evaluation scores from a reasoning model (Distill-R1) as a discriminator.}
	\label{fig:overview1}
\end{figure}

\noindent \textbf{Reasoning Model:} 
Unlike the non-reasoning models, Distill-R1 utilizes reasoning to obtain a conclusion, and thus obtaining a soft score is not straightforward. For this, we introduce a novel approach: first, we  prompt the reasoning model to think and output the final answer in a specific key-value format (JSON). In particular, we prompt the model with the instruction \texttt{``Determine whether the SQL query is correct or not for the given question. Provide final answer in JSON format with the key `correct'.''} An example prompt is provided in the Appendix (Table \ref{tab:gen_prompt_text2sql}). After the token generation process, the JSON format is identified and parsed to obtain the value for key \texttt{`correct'} along with its corresponding logit. Then, we normalize the logit via the softmax function to compute the probability $\rho$ for the possible key values (`true' or `false') and assign the score as $\rho$ for `true' or $1-\rho$ for `false'. (Figure \ref{fig:overview1}). If both `true' and `false' values are missing, a default score  $-0.5$ is returned.

\subsubsection{Planning Method}
We adopt re-ranking, a simple yet effective planning strategy where the generator first samples a set of candidate queries. These candidates are then individually scored by the discriminator, and the highest-scoring query is returned as the final output. Despite its simplicity, this planning method is widely adopted in code generation \cite{pmlr-v202-ni23b-lever} and mathematical reasoning tasks \cite{wang2023selfconsistency, li-etal-2023-making}.

In our experiments, we utilize the discriminator to score the candidates in two configurations \cite{main}:
\begin{enumerate}[topsep=0pt, partopsep=0pt, itemsep=0pt ,leftmargin=*]
	\item \textit{Naive Discriminator}: In this case, the discriminator evaluates the candidates without any additional information.
	\item \textit{Enhanced with Executability Check}: Here, the candidate SQL queries are tested on an environment to see whether they are executable (run without producing errors). The discriminator only ranks the executable candidates.
\end{enumerate}

\subsection{Evaluation Methods}
\subsubsection{Intrinsic Evaluation}

To assess LLMs' discrimination performance, we reuse the generation results from the oracle-simulation experiments in \cite{main}. Here, evaluation scripts were used to re-label the pre-generated SQL queries in simulated re-ranking experiments, where an oracle compares the top five execution results of the predicted and gold SQL queries using table cell overlap.
We evaluate the discrimination ability using a range of metrics to capture both binary classification and ranking performance.

\subsubsection{End-to-End Evaluation}

Unlike intrinsic evaluation, which assesses discrimination ability on a fixed set of queries, our end-to-end evaluation measures the effectiveness of LLMs in dynamically generating and ranking candidates within our planning framework. Specifically, we assess the full pipeline performance of our LLM planning framework with LLM-based generators and discriminators under the re-ranking planning strategy and determine their ability to generate and select correct queries in real-time decision-making scenarios. For consistency, we use the same 400 problem instances as in the intrinsic evaluation, enabling one-to-one correspondence across query batches.

\subsection{Performance Metrics}

\subsubsection{Intrinsic Evaluation}

To measure the discrimination abilities of LLMs, we employ four intrinsic metrics that evaluate both pairwise judgment and ranking quality \cite{main}. Pairwise Discrimination Accuracy \textbf{(PwAcc)} measures the percentage of times the model assigns a higher score to a correct query over an incorrect one, given a pair of queries \cite{bai2022training, touvron2023llama2}. Classification Macro F1 \textbf{(F1)} treats ``correct'' and ``incorrect'' as separate classes and computes the macro-averaged F1 score across them. Hit@1 \textbf{(H@1)} evaluates how often the top-ranked query from a candidate set is correct. Finally, mean Reciprocal Rank \textbf{(MRR)} computes the average reciprocal rank of the highest-scoring candidate in each batch.

\subsubsection{End-to-End Evaluation} 
We evaluate the end-to-end performance of the models in the LLM-based planning framework using both execution and match-based metrics. This includes execution accuracy for text-to-SQL generation, exact match accuracy to the reference gold query, and partial match performance measured by accuracy, recall, and F1 score. We utilize the semantic evaluation test suite \cite{ruiqi20} for the metric calculation.

\subsubsection{Reasoning Quality} 
To measure the richness and non-redundancy of the model’s reasoning output, we assess its reasoning quality using a set of diversity and repetition-based metrics: \textbf{Repetition Ratio} quantifies the proportion of repeated sentences, $\boldsymbol{1 - }$\textbf{TTR} (One minus Type-Token Ratio) calculates lexical simplicity by capturing the proportion of repeated sentences relative to the total, with higher values indicating greater redundancy and simpler language, \textbf{Repeated n-grams} captures the frequency of recurring n-grams (e.g., 3-grams)  by calculating the repetition ratio of phrases, \textbf{Entropy} reflects the distribution of sentence occurrences, with higher values signifying richer output. Finally, \textbf{Diversity} measures the overall variety in the generated content by utilizing sentence embeddings and cosine similarity to assess the average dissimilarity between sentence pairs. Higher diversity corresponds to lower average similarity.

\section{Results}\label{sec:results}

\subsection{Reasoning Model as Discriminator}
We investigate how our small reasoning model (Distill-R1) performs as an evaluator in our agentic LLM planning framework compared to other non-reasoning models.

\subsubsection{Distill-R1 Outperforms Larger Non-Reasoning LLMs}

\noindent \textbf{Intrinsic Evaluation.} 
Our intrinsic evaluation reveals several key takeaways. First, Distill-R1 achieves the highest classification F1 scores under both naive and executability-check settings, outperforming the non-reasoning models by approximately $58\%$ and $87\%$, respectively (Table \ref{tab:cnn_model_summary}). Second, in terms of Hit@1 and MRR, Distill-R1 exceeds all models except CodeLlama-13B. In particular, compared to larger non-reasoning LLMs such as StableCode-3B and CodeLlama-7B, Distill-R1 demonstrates improvements of roughly $4.6\%$ in Hit@1 and $2.1–2.6\%$ in MRR under the naive setting, with gains of around $3\%$ in Hit@1 and $1.4–1.9\%$ in MRR when executability checks are applied (Table \ref{tab:cnn_model_summary}). Finally, for pairwise discrimination accuracy, Distill-R1 remains highly competitive, with its performance surpassed only by StableCode-3B and CodeLlama-13B. In the naive setting, its accuracy is within $1\%$ of StableCode-3B and approximately $12\%$ lower than CodeLlama-13B. With executability checks, it is less than $0.5\%$ lower than StableCode-3B and $1.8\%$ lower than CodeLlama-13B. These findings show that, while CodeLlama-13B may lead in overall accuracy, the reasoning capabilities of Distill-R1 yield substantial improvements as a discriminator, surpassing larger non-reasoning LLMs.

\begin{table*}[htbp!]
	\small
	\centering
	\renewcommand{\arraystretch}{1.2}
	\setlength{\tabcolsep}{6pt}
	\begin{tabular}{lcccccccc}
		\toprule
		\multirow{2}{*}{\bf Models} & \multicolumn{4}{c}{\textbf{Naive Discriminator}} & \multicolumn{4}{c}{\textbf{+ Executability Check}} \\
		\cmidrule(lr){2-5} \cmidrule(lr){6-9}
		& {PwAcc} & {F1} & {H@1} & {MRR} & {PwAcc} & {F1} & {H@1} & {MRR} \\
		\midrule
		TinyLlama-1.1B & 49.39 & 37.09 & 54.75 & 61.87 & 77.75 & 37.09 & 62.00 & 66.31 \\
		Deepseek Coder-1.3B & 50.86 & 37.09 & 54.50 & 61.42 & 76.28 & 37.09 & 61.75 & 66.13 \\
		StableCode-3B & 51.59 & 37.09 & 54.00 & 61.54 & 78.73 & 37.09 & 61.25 & 66.15 \\
		CodeLlama-7B & 49.88 & 37.09 & 54.00 & 61.26 & 75.79 & 37.09 & 61.50 & 65.86 \\
		\rowcolor{green!20} 
		Distill R1-1.5B& 51.34 & \textbf{58.70} & 56.50 & 62.83 & 78.48 & \textbf{69.29} & 63.25 & 67.09 \\
		CodeLlama-13B & \textbf{58.44} & 37.09 &\textbf{ 58.50} & \textbf{64.07} & \textbf{79.95} & 37.09 & \textbf{64.75} & \textbf{67.77} \\
		%\midrule
		% \rowcolor{green!20} 
		% Deepseek R1-1.5B & 51.34 & 58.70 & 56.50 & 62.83 & 78.48 & 69.29 & 63.25 & 67.09 \\
		%\rowcolor{green!20} 
		% Deepseek R1-1.5B (hard) & 34.47 & 58.96 & 57.50 & 63.10 & 69.44 & 69.48 & 63.75 & 67.16 \\
		%	\midrule
		%	GPT-3.5-Turbo & 67.0 & 47.3 & 59.0 & 64.3 & -& - & - & - \\
		%	GPT-4-Turbo & \textbf{76.5} & 54.9 & 63.0 & 66.7 & - & - & - & - \\
		\bottomrule
	\end{tabular}
	%\vspace{-6pt}
	\caption{\label{tab:cnn_model_summary}
		Performance comparison of LLMs as discriminators during intrinsic evaluation. \textbf{Best performance} is in bold. The reasoning model is in green.
	}
	%\vspace{-11pt}
\end{table*}

\begin{figure}[!htbp]
	\centering
	\includegraphics[width=14cm,height=10.5cm,clip,keepaspectratio]{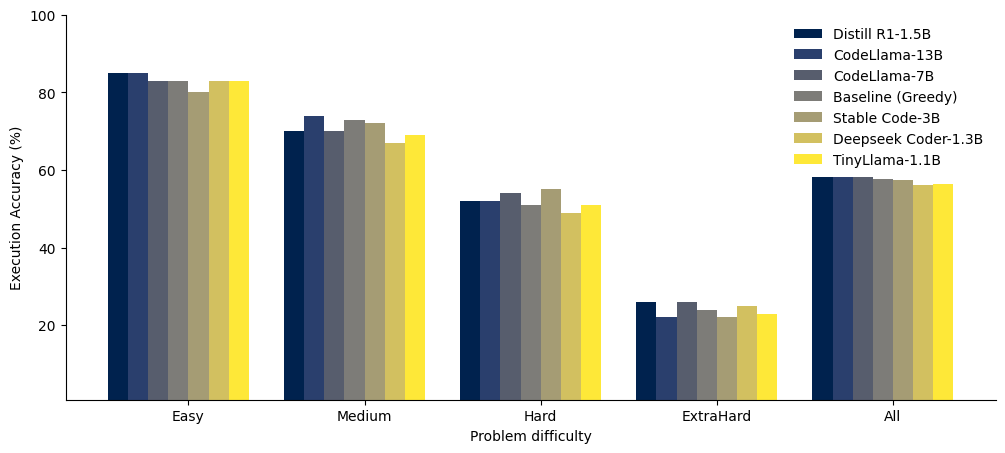}
	\caption{Performance comparison of LLMs as discriminators in terms of execution accuracy during end-to-end evaluation (with Executability Check). }
	\label{AF1}
\end{figure}

\noindent \textbf{End-To-End evaluation.}
In the end-to-end setting, Distill-R1 demonstrates superior overall execution accuracy in the naive discriminator mode, outperforming CodeLlama-13B by $3.7\%$ and achieving the highest accuracy across all problem categories except the extra hard set (Table \ref{tab:eval_results2}). When executability checks are applied, Distill-R1 maintains the top overall accuracy (Figure \ref{AF1}), and notably, its performance on extra hard problems improves  $18.2\%$ relative to CodeLlama-13B (Table \ref{tab:eval_results}). These results indicate that incorporating executability checks can significantly benefit reasoning models on more challenging queries. Meanwhile, the exact match and partial match metrics present mixed outcomes, with Distill-R1 consistently trailing CodeLlama-13B by $10–12\%$.

\begin{table*}[!htbp]
	\small
	\centering
	\renewcommand{\arraystretch}{1.2}
	\setlength{\tabcolsep}{4pt}
	\begin{tabular}{lccccccccc}
		\toprule
		\multirow{2}{*}{\bf Models} & \multicolumn{5}{c}{\textbf{Execution Accuracy}} & \multicolumn{1}{c}{\textbf{Exact Match}} & \multicolumn{3}{c}{\textbf{Partial Match}} \\
		\cmidrule(lr){2-6} \cmidrule(lr){7-7} \cmidrule(lr){8-10}
		& {Easy} & {Mid} & {Hard} & {Extra} & {All} & {Acc} & {Acc} & {Recall} & {F1} \\
		\midrule
		TinyLlama-1.1B & 80 & 66 & 46 & \textbf{21} & 53.2 & 30.00 & 66.83 & 43.69 & 50.96 \\
		Deepseek Coder-1.3B & 81 & 65 & 46 & 18 & 52.5 & 29.20 & \textbf{72.30} & 42.71 & 50.73 \\
		Stable Code-3B & 80 & 70 & 49 & 16 & 53.7 & 28.00 & 66.27 & 40.81 & 48.94 \\
		CodeLlama-7B & 80 & 64 & 44 & 19 & 51.7 & 28.20 & 65.67 & 41.19 & 49.23 \\
		CodeLlama-13B & 84 & 70 & 43 & 19 & 54.0 & \textbf{33.00} & 70.67 & \textbf{45.39} & \textbf{52.98} \\
		\rowcolor{green!20} 
		Distill R1-1.5B & \textbf{85} & \textbf{71} & \textbf{50} & 18 & \textbf{56.0} & 29.50 & 65.51 & 43.06 & 50.57 \\
		\bottomrule
	\end{tabular}
	%\vspace{-6pt}
	\caption{\label{tab:eval_results2}
		Execution accuracy and gold SQL matching performance of LLMs as discriminators in end-to-end evaluation (Naive Discriminator). \textbf{Best performance} is in bold. The reasoning model (Distill-R1) is in green.
	}
	%\vspace{-11pt}
\end{table*}

\begin{table*}[!htbp]
	\small
	\centering
	\renewcommand{\arraystretch}{1.2}
	\setlength{\tabcolsep}{4pt}
	\begin{tabular}{lccccccccc}
		\toprule
		\multirow{2}{*}{\bf Models} & \multicolumn{5}{c}{\textbf{Execution Accuracy}} & \multicolumn{1}{c}{\textbf{Exact match}} & \multicolumn{3}{c}{\textbf{Partial match}} \\
		\cmidrule(lr){2-6} \cmidrule(lr){7-7} \cmidrule(lr){8-10}
		& {Easy} & {Mid} & {Hard} & {Extra} & {All} &{Acc} & {Acc} & {Recall} & {F1} \\
		\midrule
		TinyLlama-1.1B & 83 & 69 & 51 & 23 & 56.5  & 31.20 & 65.95 & 46.56 & 53.00 \\
		Deepseek Coder-1.3B & 83 & 67 & 49 & 25 & 56 & 29.70 & \textbf{68.72} & 44.84 & 52.16 \\
		Stable Code-3B& 80 & 72 & \textbf{55} & 22 & 57.3 & 28.70 & 63.94 & 42.37 & 49.71 \\
		CodeLlama-7B& 83 & 70 & 54 & \textbf{26} & \textbf{58.3} & 30.20 & 65.80 & 44.43 & 51.69 \\
		CodeLlama-13B & 85 & \textbf{74} & 52 & 22 & \textbf{58.3} & \textbf{34.50} & 67.50 & \textbf{47.08} & \textbf{53.68} \\
		\rowcolor{green!20} 
		Distill R1-1.5B & \textbf{85} & 70 & 52 & \textbf{26} & \textbf{58.3} & 30.20 & 63.73 & 45.45 & 52.09 \\
		\bottomrule
	\end{tabular}
	%\vspace{-6pt}
	\caption{\label{tab:eval_results}
		Execution accuracy and gold SQL matching performance of LLMs as discriminators in end-to-end evaluation (+ Executability Check). \textbf{Best performance} is in bold. The reasoning model (Distill-R1) is in green.
	}
	%\vspace{-11pt}
\end{table*}

\subsubsection{Increasing Test-Time Compute Yields Diminishing Returns}
Given the strong performance of Distill-R1 as a discriminator in both intrinsic and end-to-end evaluations, we further investigate the impact of test-time compute budget on its discrimination capabilities. Specifically, we explore whether allocating additional computation for reasoning enhances its performance. Experiments are conducted using Distill-R1 as the discriminator under varying token budgets.

\begin{figure}[!htbp]
	\centering
	\includegraphics[width=14cm,height=10.5cm,clip,keepaspectratio]{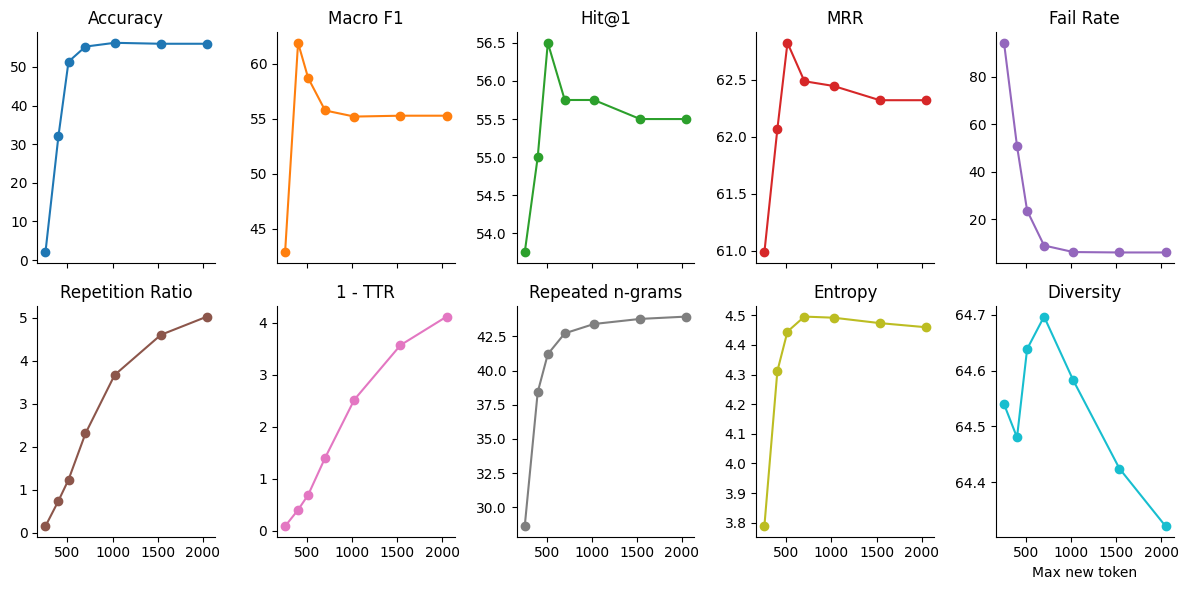}
	\caption{Discrimination performance (\textit{top row}) and reasoning quality (\textit{bottom row}) of the reasoning model (Distill-R1) as a function of test-time compute budget (token limit).  }
	\label{AF0}
\end{figure}  

Our results show that increasing test-time compute (via the maximum new token limit) significantly enhances discrimination performance initially, but levels off after reaching a certain threshold. At very low token limits (e.g., 256 tokens), the reasoning process is heavily constrained, which causes drastically low pairwise discrimination accuracy ($<2\%$) and a high failure rate ($94+\%$ ; Figure \ref{AF0}; Appendix: Table \ref{tab:intrin_eval}). As the token limit increases, the pairwise accuracy improves gradually and peaks around 1024 tokens. Beyond this point, additional tokens provide only marginal gains, with accuracy increasing by less than $0.4\%$ from 1024 to 2048 tokens. Similar trends are observed in other metrics, such as F1, Hit@1, and MRR, while the failure rate decreases sharply with more tokens before stabilizing.

In terms of reasoning quality, our results suggest diminishing returns with increased test-time compute. As the maximum token limit increases, $1 - $TTR, repetition ratio, and repeated n-grams all rise (Figure \ref{AF0}; Appendix: Table \ref{tab:intrin_eval}), indicating that the model is producing more lexically redundant and repetitive outputs, rather than generating novel content. This suggests that beyond a certain point, additional compute budget leads to more verbose but not necessarily more insightful reasoning. Furthermore, entropy and diversity metrics exhibit peaks between 512 and 1024 tokens and then plateau or slightly decline (Appendix: Table \ref{tab:intrin_eval}). These indicate that the model's reasoning variety and information richness improve only up to a moderate compute budget, after which the quality and variety of reasoning stop improving, with longer outputs becoming more repetitive rather than more insightful. Taken together, these results show that while a minimal token threshold is necessary to enable coherent reasoning, simply increasing token limits beyond this threshold leads to diminishing improvements in both discrimination performance and reasoning quality.

\subsubsection{Additional Context Does Not Guarantee Performance Gains }
Next, we explore whether adding more context to the prompt helps the reasoning process and improves the discrimination performance of Distill-R1. For this, we add the corresponding database tables to the prompt as context (Appendix: Table \ref{tab:gen_prompt_text2sql2}) and compare the discrimination performance with the case where we don't add schema tables as context (Appendix: Table \ref{tab:gen_prompt_text2sql}). We set the maximum token limit to 1024 for both cases. 

\begin{table*}[!htbp]
	\small  % or \footnotesize or \scriptsize for smaller font
	\centering
	\renewcommand{\arraystretch}{1.2}
	\setlength{\tabcolsep}{4pt}
	\begin{tabular}{lcccccccccc}
		\toprule
		\multirow{2}{*}{\bf Prompt}& \multicolumn{5}{c}{\textbf{Discriminator Performance}} & \multicolumn{5}{c}{\textbf{Reasoning Quality}} \\
		\cmidrule(lr){2-6} \cmidrule(lr){7-11}
		{\bf Condition} & \multirow{2}{*}{{PwAcc}} & \multirow{2}{*}{{F1}} & \multirow{2}{*}{{H@1}} & \multirow{2}{*}{{MRR}} & {Fail} & {Rep.} & \multirow{2}{*}{{1-TTR}} & {Rep.} & \multirow{2}{*}{{Entropy}} & \multirow{2}{*}{{Diversity}} \\
		& & & & & {Rate} & {Ratio} & &{n-grams} & & \\
		\midrule
		w/o Schema  & 56.23 & 55.20 & 55.75 & 62.45 & 6.10 & 3.67 & 2.52 & 43.40 & 4.49 & 64.58 \\
		with Schema & 53.79 & 53.90 & 56.25 & 62.56 & 19.24 & 7.07 & 4.94 & 48.73 & 4.68 & 64.14 \\
		\midrule
		Diff. (\%)     & -4.34  & -2.36  & +0.90  & +0.18  & +215.41& +92.64  & +96.03  & +12.28  & +4.23  & -0.68 \\
		\bottomrule
	\end{tabular}
	%\vspace{-6pt}
	\caption{\label{tab:schema}
		Discrimination and reasoning performance under different prompt conditions for Distill-R1 (Naive Discriminator setting).	}
	%\vspace{-11pt}
\end{table*}

Our intrinsic evaluation results indicate that providing schema information as context has little to no benefit. Compared to the baseline without schema, the overall discrimination accuracy and F1 score drop by approximately $4.3\%$ and $2.4\%$, respectively, while improvements in Hit@1 ($0.9\%$) and MRR ($0.18\%$) are negligible (Table \ref{tab:schema}). Notably, the failure rate increases dramatically by over 215\%, suggesting that the additional context may actually hinder the model’s ability to evaluate queries. In terms of reasoning quality, repetition ratio and repeated n-grams increase drastically by $90+\%$ while $1-$TTR shows a $12\%$ rise with schema information added to the prompt. These indicate a higher degree of repetition and lexical redundancy in the reasoning. The changes in entropy and diversity are minimal. These findings suggest that augmenting the prompt with the database schema does not yield significant improvements in discrimination performance and may introduce more redundancy and instability in the reasoning process.

\subsubsection{Minimal Effect of Logits on Ranking Performance}

Finally, we assess the impact of using logits for ranking on the discrimination performance of the reasoning model. For this, we calculate the final discrimination score using the reasoning text (binary score based on the true/false decision) instead of the logit-based soft discrimination score (Figure \ref{fig:overview1}). 

\begin{table*}[!htbp]
	\small  % or \footnotesize or \scriptsize for smaller font
	\centering
	\renewcommand{\arraystretch}{1.2}
	\setlength{\tabcolsep}{6pt}
	\begin{tabular}{lcccccccc}
		\toprule
		\multirow{2}{*}{\bf Condition} & \multicolumn{4}{c}{\textbf{Naive Discriminator}} & \multicolumn{4}{c}{\textbf{+ Executability Check}} \\
		\cmidrule(lr){2-5} \cmidrule(lr){6-9}
		& \textbf{PwAcc} & \textbf{F1} & \textbf{H@1} & \textbf{MRR} & \textbf{PwAcc} & \textbf{F1} & \textbf{H@1} & \textbf{MRR} \\
		\midrule
		with Logits & 51.34 & 58.70 & 56.50 & 62.83 & 78.48 & 69.29 & 63.25 & 67.09 \\
		w/o Logits & 34.47 & 58.96 & 57.50 & 63.10 & 69.44 & 69.48 & 63.75 & 67.16 \\
		\midrule
		Diff. (\%) & -32.84 & +0.44 & +1.77 & +0.43 & -11.53 & +0.27 & +0.79 & +0.10 \\
		\bottomrule
	\end{tabular}
	%\vspace{-6pt}
	\caption{\label{tab:cnn_model_summary3}
		Performance of intrinsic evaluation with and w/o logits for Distill-R1.	}
	%\vspace{-11pt}
\end{table*}

\begin{table*}[!htbp]
	\small
	\centering
	\renewcommand{\arraystretch}{1.2}
	\setlength{\tabcolsep}{4pt}
	\begin{tabular}{lcccccccccc}
		\toprule
		\multirow{2}{*}{\bf Condition} & \multicolumn{5}{c}{\textbf{Naive Discriminator}} & \multicolumn{5}{c}{\textbf{+ Executability Check}} \\
		\cmidrule(lr){2-6} \cmidrule(lr){7-11}
		& \textbf{Easy} & \textbf{Mid} & \textbf{Hard} & \textbf{Extra} & \textbf{All} & \textbf{Easy} & \textbf{Mid} & \textbf{Hard} & \textbf{Extra} & \textbf{All} \\
		\midrule
		with Logits  & 85.00 & 71.00 & 50.00 & 18.00 & 56.00 & 85.00 & 70.00 & 52.00 & 26.00 & 58.30 \\
		w/o Logits   & 86.00 & 67.00 & 50.00 & 18.00 & 55.20 & 86.00 & 70.00 & 51.00 & 25.00 & 58.00 \\
		\midrule
		Diff (\%)      & +1.18 & -5.63 & 0.00 & 0.00 & -1.43 & +1.18 & 0.00 & -1.92 & -3.85 & -0.51 \\
		\bottomrule
	\end{tabular}
	%\vspace{-6pt}
	\caption{\label{tab:cnn_model_summary4}
		Execution accuracy in end-to-end evaluation with and w/o logits for Distill-R1.
	}
	%\vspace{-11pt}
\end{table*}

In intrinsic evaluation, switching from a logit-based soft score to a binary score (without logits) results in a substantial drop in overall pairwise discrimination accuracy ($32.8\%$ lower in naive setting and $11.5\%$ lower with executability checks; Table \ref{tab:cnn_model_summary3}) while F1, Hit@1, and MRR metrics remained largely unchanged. In contrast, the end-to-end evaluation reveals only a marginal reduction in overall execution accuracy ($< 1.5\%$ ; Table \ref{tab:cnn_model_summary4}). These findings indicate that the final discrimination scoring method (with or without logits) has little impact on overall task performance in practical LLM planning scenarios.

\subsection{Reasoning Model as Generator}
\begin{figure}[!htbp]
	\centering
	\includegraphics[width=13.5cm,height=10.5cm,clip,keepaspectratio]{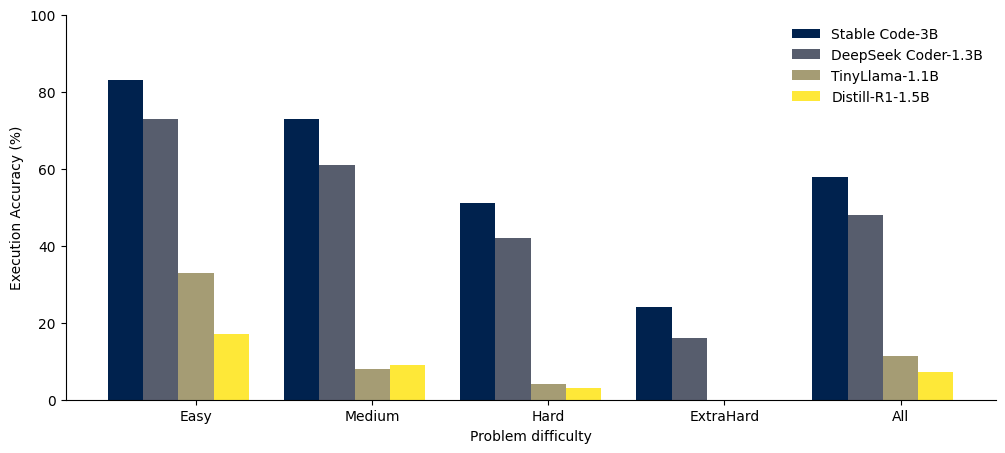}
	\caption{Performance comparison of LLMs as generators in terms of execution accuracy.}
	\label{AF2}
\end{figure}

Our results show that the reasoning model, Distill-R1, is not an effective generator. In particular, with Distill-R1 as the baseline, the smallest non-reasoning model (TinyLlama-1.1B) shows improvements of approximately $56.9\%$ in overall execution accuracy, $87.5\%$ in exact match accuracy, and $49.4\%$ in partial match F1 score (Table \ref{tab:merged} and Figure \ref{AF2}).

\begin{table*}[!htbp]
	\small
	\centering
	\renewcommand{\arraystretch}{1.2}
	\setlength{\tabcolsep}{4pt}
	\begin{tabular}{lccccccccc}
		\toprule
		\multirow{2}{*}{\bf Models} & \multicolumn{5}{c}{\textbf{Execution Accuracy}} & \multicolumn{1}{c}{\textbf{Exact match}} & \multicolumn{3}{c}{\textbf{Partial match}} \\
		\cmidrule(lr){2-6} \cmidrule(lr){7-7} \cmidrule(lr){8-10}
		& {Easy} & {Mid} & {Hard} & {Extra} & {All} &{Acc} & {Acc} & {Recall} & {F1} \\
		\midrule
		\rowcolor{green!20} Distill-R1-1.5B & 17.00 & 9.00 & 3.00 & 0.00 & 7.20 & 4.00 & 53.87 & 14.23 & 27.22 \\
		TinyLlama-1.1B & 33.00 & 8.00 & 4.00 & 0.00 & 11.30 & 7.50 & 36.28 & 17.04 & 40.66 \\
		DeepSeek Coder-1.3B & 73.00 & 61.00 & 42.00 & 16.00 & 48.00 & 19.80 & 66.11 & 29.29 & 37.81 \\
		Stable Code-3B & \textbf{83.00} & \textbf{73.00} & \textbf{51.00} & \textbf{24.00} & \textbf{57.80} & \textbf{33.80} & \textbf{69.50} & \textbf{44.70} & \textbf{52.36} \\
		\bottomrule
	\end{tabular}
	%\vspace{-6pt}
	\caption{\label{tab:merged}
		Execution accuracy and gold SQL matching performance in end-to-end evaluation as a generator (Greedy). \textbf{Best performance} is in bold.
	}
	%\vspace{-11pt}
\end{table*}

Moreover, performance among non-reasoning models increases with model size. A mid-sized model yields gains of roughly $566.7\%$ in execution accuracy, $395\%$ in exact match accuracy, and $38.9\%$ in partial match F1 score relative to Distill-R1. The largest model further enhances these metrics with improvements of $702.8\%$, $745\%$, and $92.3\%$ in execution accuracy, exact match accuracy, and partial match F1 score, respectively. These findings underscore two key points: first, that the reasoning model is outperformed by even smaller non-reasoning models, and second, that scaling non-reasoning architectures leads to progressively higher performance gains in SQL generation tasks.

\section{Discussion}\label{sec:limitation}

In this work, we study the role of a reasoning model within the generator-discriminator LLM planning framework\cite{main}. In particular, we use the re-ranking planning method and benchmark a reasoning model (Distill-R1-1.5B) against several non-reasoning LLMs in the text-to-SQL task. Our investigation reveals several key insights into the roles of reasoning models in LLM planning frameworks, with significant implications for both the design and deployment of such models in an agentic framework.

\subsection{Distill-R1 Outperforms Larger Non-Reasoning LLMs as Discriminator}
First, our findings show that the reasoning model excels as a discriminator not only in distinguishing correct from incorrect queries but also by achieving the highest overall execution accuracy in the LLM planning framework,  outperforming larger non-reasoning models in both cases.
During our intrinsic evaluation, Distill-R1 shows strong capabilities in discriminating between correct and incorrect queries more effectively than its non-reasoning counterparts. In particular, compared to CodeLlama-7B, Distill-R1 demonstrates improvements of $2.9\%$ in pairwise accuracy, $58\%$ in F1, $4.6\%$ in Hit@1, and $2.6\%$ in MRR under the naive setting, which become $3.5\%$, $87\%$, $2.8\%$, and $1.9\%$ respectively, when executability checks are applied (Table \ref{tab:cnn_model_summary}). 
In terms of overall execution accuracy during end-to-end evaluation of our LLM planning framework, \textit{Distill-R1 with only 1.5B parameters outperformed non-reasoning models that are more than 8 times larger} ($3.7\%$ improvement compared to CodeLlama-13B; Table \ref{tab:eval_results2}). These findings suggest that reasoning models are particularly effective in evaluating candidate queries, even when a smaller model is used. Thus, by leveraging their inherent reasoning abilities, such models can discern subtle differences between candidates more effectively than non-reasoning models. 
Moreover, the performance difference between logit-based soft scoring and binary true/false discrimination is minimal ($<1.5\%$) for Distill-R1, showing that reasoning models are good discriminators in the LLM planning framework even when only binary outcomes are available. This can provide a significant advantage in commercial settings where access to logit values for reasoning models may be restricted or expensive. Together, these results suggest that reasoning models hold substantial promise for robust discrimination tasks, and further research could drive even greater performance improvements.

\subsection{Inherent Limits of Reasoning Capabilities }
Second, our findings demonstrate that for reasoning models, there is a limit to the benefits of increased test-time compute for evaluating candidates. A very low token limit severely restricts the reasoning process, leading to extremely low accuracy and a very high failure rate, indicating the necessity of a minimum compute threshold. As the token budget increases, performance improves and peaks around 1024 tokens, beyond which the gains become negligible. These show that allocating more test-time compute budget does not directly translate into more insightful reasoning. We also observe that increasing the token budget leads to more redundancy and repetition in reasoning outputs  (Figure \ref{AF0}; Appendix: Table \ref{tab:intrin_eval}), indicating the occurrence of circular reasoning and loss of context. Furthermore, we find that incorporating database schema information into the prompt does not offer significant improvements for Distill-R1 (Table \ref{tab:schema}). In fact, it can introduce redundancy and even elevate failure rates. Our findings are consistent with earlier studies, which also showed a minimum computing budget, a diminishing return after a certain threshold, and an inverse relationship between CoT length and accuracy \cite{time-scale-o1, wu2025lessunderstandingchainofthoughtlength}.
Together, these results indicate that \textit{there is a limit to the logical capabilities of these reasoning models, and providing more context or allowing more compute budget for reasoning is not sufficient for improving their performance}. Future studies should focus not only on scaling compute and maximizing context window but also on targeted fine-tuning of such reasoning models.

\subsection{Generation is Harder than Discrimination} 
Third, while Distill-R1 shows strong performance as a discriminator, it performs poorly as a generator. In fact, for generating high-quality SQL queries, it is outperformed by smaller non-reasoning models (Table \ref{tab:merged} and Figure \ref{AF2}). This suggests that while reasoning models are effective at evaluating candidate queries, they may not be well-suited for generating them. This result is not surprising, as SQL generation from natural language involves complex reasoning that can branch into convoluted CoT. With our earlier results indicating a fundamental limitation in the reasoning capabilities, navigating through a complex CoT is challenging for reasoning models. Hence, \textit{while previous studies have shown that for non-reasoning LLMs, the discriminator role is typically more challenging and significant than the generator role \cite{main, west2024paradox, huang2024llm_selfcorrect, wang-etal-2023-chatgpt-defend}, our results indicate the opposite for reasoning models} , which is similar to humans \cite{gu-etal-2023-dont}.

\subsection{Future Works and Limitations}
In this study, we demonstrate that reasoning models can be employed to generate soft scores for evaluating candidates, which is a non-trivial task given that CoT reasoning may produce arbitrarily long token outputs, complicating soft score extraction. By leveraging our approach, it is potentially possible to fine-tune such models (with soft score as the cost function) and adjust their reasoning approaches to improve performance even further. We plan to explore this in future studies.

Our study has several limitations. The experiments are conducted within the specific context of text-to-SQL generation and discrimination, which may not directly generalize to other domains or reasoning tasks. Additionally, while our analysis provides insights into the trade-offs between compute budget and performance, further research is necessary to fully understand the interplay between model architecture, token limits, and task complexity. Lastly, we use a relatively small reasoning model (Distill-R1, 1.5B parameters) and a reduced dataset due to our limited budget for computational resources. However, our goal in this work is to investigate and gain a deeper understanding of reasoning models within LLM planning frameworks.  In this aspect, we believe our findings will provide significant insights even for larger reasoning models.

\section{Conclusion}\label{sec:conclude}
In this work, we explore the role of a reasoning model, Distilled DeepSeek-R1-1.5B, within an LLM planning framework for text-to-SQL task. We introduce a novel approach to extract soft scores from its CoT outputs for discrimination and systematically evaluate its capabilities as a discriminator well as a generator.  
We show that DeepSeek-R1 outperforms much larger non-reasoning models in query discrimination, highlighting the potential of such models as discriminators within LLM planning frameworks. Our study also reveals that reasoning models face inherent limitations. In particular, increasing test-time compute and additional context yield diminishing returns. Finally, while DeepSeek-R1 excels at evaluating candidate queries, it struggles as a generator, where even smaller non-reasoning models deliver higher-quality SQL outputs. 
Though our experiments are constrained in scope and scale, these insights can point to promising directions for developing smaller, more efficient agentic systems that strategically integrate reasoning models into planning frameworks.

\subsection*{Data and Code Availability}
All code, datasets, and detailed instructions necessary to reproduce our results are publicly available at: \href{https://github.com/MDFahimAnjum/llm-planning-with-reasoning}{https://github.com/MDFahimAnjum/llm-planning-with-reasoning}.

\subsection*{Computing Hardware} 
All experiments are conducted on commercially available, consumer-grade laptops equipped with NVIDIA GeForce RTX 4080 (12GB) and RTX 3070 (8GB) GPUs.

%\subsubsection*{Acknowledgments}
%Use unnumbered third level headings for the acknowledgments. All acknowledgments go at the end of the paper. Do not include acknowledgments in the anonymized submission, only in the final paper.

\bibliographystyle{plain} % Use plain style for numbered citations
\setlength{\itemindent}{0pt} % Set the first line indent to 0
\bibliography{custom.bib} % for local

\begin{thebibliography}{10}

\bibitem{bai2022training}
Yuntao Bai, Andy Jones, Kamal Ndousse, Amanda Askell, Anna Chen, Nova DasSarma, Dawn Drain, Stanislav Fort, Deep Ganguli, Tom Henighan, Nicholas Joseph, Saurav Kadavath, Jackson Kernion, Tom Conerly, Sheer El-Showk, Nelson Elhage, Zac Hatfield-Dodds, Danny Hernandez, Tristan Hume, Scott Johnston, Shauna Kravec, Liane Lovitt, Neel Nanda, Catherine Olsson, Dario Amodei, Tom Brown, Jack Clark, Sam McCandlish, Chris Olah, Ben Mann, and Jared Kaplan.
\newblock Training a helpful and harmless assistant with reinforcement learning from human feedback, 2022.

\bibitem{balachandran2025inferencetimescalingcomplextasks}
Vidhisha Balachandran, Jingya Chen, Lingjiao Chen, Shivam Garg, Neel Joshi, Yash Lara, John Langford, Besmira Nushi, Vibhav Vineet, Yue Wu, and Safoora Yousefi.
\newblock Inference-time scaling for complex tasks: Where we stand and what lies ahead, 2025.

\bibitem{chen2025reasoningerasurveylong}
Qiguang Chen, Libo Qin, Jinhao Liu, Dengyun Peng, Jiannan Guan, Peng Wang, Mengkang Hu, Yuhang Zhou, Te~Gao, and Wanxiang Che.
\newblock Towards reasoning era: A survey of long chain-of-thought for reasoning large language models, 2025.

\bibitem{chen2024unlocking}
Qiguang Chen, Libo Qin, Jiaqi WANG, Jingxuan Zhou, and Wanxiang Che.
\newblock Unlocking the capabilities of thought: A reasoning boundary framework to quantify and optimize chain-of-thought.
\newblock In {\em The Thirty-eighth Annual Conference on Neural Information Processing Systems}, 2024.

\bibitem{main}
Ziru Chen, Michael White, Raymond Mooney, Ali Payani, Yu~Su, and Huan Sun.
\newblock When is tree search useful for llm planning? it depends on the discriminator, 2024.

\bibitem{deepseekai2025deepseekr1incentivizingreasoningcapability}
DeepSeek-AI.
\newblock Deepseek-r1: Incentivizing reasoning capability in llms via reinforcement learning, 2025.

\bibitem{feng2024alphazerolike}
Xidong Feng, Ziyu Wan, Muning Wen, Stephen~Marcus McAleer, Ying Wen, Weinan Zhang, and Jun Wang.
\newblock Alphazero-like tree-search can guide large language model decoding and training, 2024.

\bibitem{google2024gemini}
Google.
\newblock Gemini 2.0 flash thinking experimental.
\newblock \url{https://deepmind.google/technologies/gemini/flash-thinking/}, December 2024.
\newblock Accessed: April 2025.

\bibitem{gu-etal-2023-dont}
Yu~Gu, Xiang Deng, and Yu~Su.
\newblock Don{'}t generate, discriminate: A proposal for grounding language models to real-world environments.
\newblock In Anna Rogers, Jordan Boyd-Graber, and Naoaki Okazaki, editors, {\em Proceedings of the 61st Annual Meeting of the Association for Computational Linguistics (Volume 1: Long Papers)}, pages 4928--4949, Toronto, Canada, July 2023. Association for Computational Linguistics.

\bibitem{huang2024llm_selfcorrect}
Jie Huang, Xinyun Chen, Swaroop Mishra, Huaixiu~Steven Zheng, Adams~Wei Yu, Xinying Song, and Denny Zhou.
\newblock Large language models cannot self-correct reasoning yet.
\newblock In {\em The Twelfth International Conference on Learning Representations}, 2024.

\bibitem{jin-etal-2024-impact}
Mingyu Jin, Qinkai Yu, Dong Shu, Haiyan Zhao, Wenyue Hua, Yanda Meng, Yongfeng Zhang, and Mengnan Du.
\newblock The impact of reasoning step length on large language models.
\newblock In Lun-Wei Ku, Andre Martins, and Vivek Srikumar, editors, {\em Findings of the Association for Computational Linguistics: ACL 2024}, pages 1830--1842, Bangkok, Thailand, August 2024. Association for Computational Linguistics.

\bibitem{kadavath2022language}
Saurav Kadavath, Tom Conerly, Amanda Askell, Tom Henighan, Dawn Drain, Ethan Perez, Nicholas Schiefer, Zac Hatfield-Dodds, Nova DasSarma, Eli Tran-Johnson, Scott Johnston, Sheer El-Showk, Andy Jones, Nelson Elhage, Tristan Hume, Anna Chen, Yuntao Bai, Sam Bowman, Stanislav Fort, Deep Ganguli, Danny Hernandez, Josh Jacobson, Jackson Kernion, Shauna Kravec, Liane Lovitt, Kamal Ndousse, Catherine Olsson, Sam Ringer, Dario Amodei, Tom Brown, Jack Clark, Nicholas Joseph, Ben Mann, Sam McCandlish, Chris Olah, and Jared Kaplan.
\newblock Language models (mostly) know what they know, 2022.

\bibitem{ke-etal-2023-decompeval}
Pei Ke, Fei Huang, Fei Mi, Yasheng Wang, Qun Liu, Xiaoyan Zhu, and Minlie Huang.
\newblock {D}ecomp{E}val: Evaluating generated texts as unsupervised decomposed question answering.
\newblock In Anna Rogers, Jordan Boyd-Graber, and Naoaki Okazaki, editors, {\em Proceedings of the 61st Annual Meeting of the Association for Computational Linguistics (Volume 1: Long Papers)}, pages 9676--9691, Toronto, Canada, July 2023. Association for Computational Linguistics.

\bibitem{li-etal-2023-making}
Yifei Li, Zeqi Lin, Shizhuo Zhang, Qiang Fu, Bei Chen, Jian-Guang Lou, and Weizhu Chen.
\newblock Making language models better reasoners with step-aware verifier.
\newblock In Anna Rogers, Jordan Boyd-Graber, and Naoaki Okazaki, editors, {\em Proceedings of the 61st Annual Meeting of the Association for Computational Linguistics (Volume 1: Long Papers)}, pages 5315--5333, Toronto, Canada, July 2023. Association for Computational Linguistics.

\bibitem{madaan2023selfrefine}
Aman Madaan, Niket Tandon, Prakhar Gupta, Skyler Hallinan, Luyu Gao, Sarah Wiegreffe, Uri Alon, Nouha Dziri, Shrimai Prabhumoye, Yiming Yang, Shashank Gupta, Bodhisattwa~Prasad Majumder, Katherine Hermann, Sean Welleck, Amir Yazdanbakhsh, and Peter Clark.
\newblock Self-refine: Iterative refinement with self-feedback.
\newblock In {\em Thirty-seventh Conference on Neural Information Processing Systems}, 2023.

\bibitem{planning_brain}
Marcelo~G. Mattar and Máté Lengyel.
\newblock Planning in the brain.
\newblock {\em Neuron}, 110(6):914--934, 2022.

\bibitem{newellsimon1956planning}
A.~Newell and H.~Simon.
\newblock The logic theory machine--a complex information processing system.
\newblock {\em IRE Transactions on Information Theory}, 2(3):61--79, 1956.

\bibitem{pmlr-v202-ni23b-lever}
Ansong Ni, Srini Iyer, Dragomir Radev, Veselin Stoyanov, Wen-Tau Yih, Sida Wang, and Xi~Victoria Lin.
\newblock {LEVER}: Learning to verify language-to-code generation with execution.
\newblock In Andreas Krause, Emma Brunskill, Kyunghyun Cho, Barbara Engelhardt, Sivan Sabato, and Jonathan Scarlett, editors, {\em Proceedings of the 40th International Conference on Machine Learning}, volume 202 of {\em Proceedings of Machine Learning Research}, pages 26106--26128. PMLR, 23--29 Jul 2023.

\bibitem{openai2024openaio1card}
OpenAI.
\newblock Openai o1 system card, 2024.

\bibitem{parashar2025inferencetimecomputationsllmreasoning}
Shubham Parashar, Blake Olson, Sambhav Khurana, Eric Li, Hongyi Ling, James Caverlee, and Shuiwang Ji.
\newblock Inference-time computations for llm reasoning and planning: A benchmark and insights, 2025.

\bibitem{plangen}
Mihir Parmar, Xin Liu, Palash Goyal, Yanfei Chen, Long Le, Swaroop Mishra, Hossein Mobahi, Jindong Gu, Zifeng Wang, Hootan Nakhost, Chitta Baral, Chen-Yu Lee, Tomas Pfister, and Hamid Palangi.
\newblock Plangen: A multi-agent framework for generating planning and reasoning trajectories for complex problem solving, 2025.

\bibitem{stable-code-3b}
Nikhil Pinnaparaju, Reshinth Adithyan, Duy Phung, Jonathan Tow, James Baicoianu, and Nathan Cooper.
\newblock Stable code 3b.

\bibitem{qwen2025qwq32b}
{Qwen Team}.
\newblock {QwQ-32B: Embracing the Power of Reinforcement Learning}.
\newblock \url{https://qwenlm.github.io/blog/qwq-32b/}, March 2025.
\newblock Accessed: April 2025.

\bibitem{codellama}
Baptiste Rozière, Jonas Gehring, Fabian Gloeckle, Sten Sootla, Itai Gat, Xiaoqing~Ellen Tan, Yossi Adi, Jingyu Liu, Romain Sauvestre, Tal Remez, Jérémy Rapin, Artyom Kozhevnikov, Ivan Evtimov, Joanna Bitton, Manish Bhatt, Cristian~Canton Ferrer, Aaron Grattafiori, Wenhan Xiong, Alexandre Défossez, Jade Copet, Faisal Azhar, Hugo Touvron, Louis Martin, Nicolas Usunier, Thomas Scialom, and Gabriel Synnaeve.
\newblock Code llama: Open foundation models for code, 2024.

\bibitem{russel2010}
Stuart Russell and Peter Norvig.
\newblock {\em Artificial Intelligence: A Modern Approach}.
\newblock Prentice Hall, 3 edition, 2010.

\bibitem{saha2025learningplanreason}
Swarnadeep Saha, Xian Li, Marjan Ghazvininejad, Jason Weston, and Tianlu Wang.
\newblock Learning to plan \& reason for evaluation with thinking-llm-as-a-judge, 2025.

\bibitem{setlur2025scalingtesttimecomputeverification}
Amrith Setlur, Nived Rajaraman, Sergey Levine, and Aviral Kumar.
\newblock Scaling test-time compute without verification or rl is suboptimal, 2025.

\bibitem{deepseek_coder_1.3b_instruct}
DeepSeek~AI Team.
\newblock Deepseek-coder-1.3b-instruct, 2024.
\newblock Fine-tuned on 2 billion tokens of instruction data, designed for advanced code generation and infilling tasks with a context length of 16K tokens.

\bibitem{distilledr1}
DeepSeek~AI Team.
\newblock Deepseek-r1-distill-qwen-1.5b: Compact reasoning model for mathematical and coding tasks, 2025.

\bibitem{touvron2023llama2}
Hugo Touvron, Louis Martin, Kevin Stone, Peter Albert, Amjad Almahairi, Yasmine Babaei, Nikolay Bashlykov, Soumya Batra, Prajjwal Bhargava, Shruti Bhosale, Dan Bikel, Lukas Blecher, Cristian~Canton Ferrer, Moya Chen, Guillem Cucurull, David Esiobu, Jude Fernandes, Jeremy Fu, Wenyin Fu, Brian Fuller, Cynthia Gao, Vedanuj Goswami, Naman Goyal, Anthony Hartshorn, Saghar Hosseini, Rui Hou, Hakan Inan, Marcin Kardas, Viktor Kerkez, Madian Khabsa, Isabel Kloumann, Artem Korenev, Punit~Singh Koura, Marie-Anne Lachaux, Thibaut Lavril, Jenya Lee, Diana Liskovich, Yinghai Lu, Yuning Mao, Xavier Martinet, Todor Mihaylov, Pushkar Mishra, Igor Molybog, Yixin Nie, Andrew Poulton, Jeremy Reizenstein, Rashi Rungta, Kalyan Saladi, Alan Schelten, Ruan Silva, Eric~Michael Smith, Ranjan Subramanian, Xiaoqing~Ellen Tan, Binh Tang, Ross Taylor, Adina Williams, Jian~Xiang Kuan, Puxin Xu, Zheng Yan, Iliyan Zarov, Yuchen Zhang, Angela Fan, Melanie Kambadur, Sharan Narang, Aurelien Rodriguez, Robert Stojnic, Sergey Edunov, and Thomas
  Scialom.
\newblock Llama 2: Open foundation and fine-tuned chat models, 2023.

\bibitem{wang-etal-2023-chatgpt-defend}
Boshi Wang, Xiang Yue, and Huan Sun.
\newblock Can {C}hat{GPT} defend its belief in truth? evaluating {LLM} reasoning via debate.
\newblock In Houda Bouamor, Juan Pino, and Kalika Bali, editors, {\em Findings of the Association for Computational Linguistics: EMNLP 2023}, pages 11865--11881, Singapore, December 2023. Association for Computational Linguistics.

\bibitem{wang2023selfconsistency}
Xuezhi Wang, Jason Wei, Dale Schuurmans, Quoc~V Le, Ed~H. Chi, Sharan Narang, Aakanksha Chowdhery, and Denny Zhou.
\newblock Self-consistency improves chain of thought reasoning in language models.
\newblock In {\em The Eleventh International Conference on Learning Representations}, 2023.

\bibitem{wang2025examiningfalsepositivesinference}
Yu~Wang, Nan Yang, Liang Wang, and Furu Wei.
\newblock Examining false positives under inference scaling for mathematical reasoning, 2025.

\bibitem{wei2023chainofthoughtpromptingelicitsreasoning}
Jason Wei, Xuezhi Wang, Dale Schuurmans, Maarten Bosma, Brian Ichter, Fei Xia, Ed~Chi, Quoc Le, and Denny Zhou.
\newblock Chain-of-thought prompting elicits reasoning in large language models, 2023.

\bibitem{wei2022cot}
Jason Wei, Xuezhi Wang, Dale Schuurmans, Maarten Bosma, brian ichter, Fei Xia, Ed~Chi, Quoc~V Le, and Denny Zhou.
\newblock Chain-of-thought prompting elicits reasoning in large language models.
\newblock In S.~Koyejo, S.~Mohamed, A.~Agarwal, D.~Belgrave, K.~Cho, and A.~Oh, editors, {\em Advances in Neural Information Processing Systems}, volume~35, pages 24824--24837. Curran Associates, Inc., 2022.

\bibitem{west2024paradox}
Peter West, Ximing Lu, Nouha Dziri, Faeze Brahman, Linjie Li, Jena~D. Hwang, Liwei Jiang, Jillian Fisher, Abhilasha Ravichander, Khyathi Chandu, Benjamin Newman, Pang~Wei Koh, Allyson Ettinger, and Yejin Choi.
\newblock The generative {AI} paradox: {\textquotedblleft}what it can create, it may not understand{\textquotedblright}.
\newblock In {\em The Twelfth International Conference on Learning Representations}, 2024.

\bibitem{wolf-etal-2020-transformers}
Thomas Wolf, Lysandre Debut, Victor Sanh, Julien Chaumond, Clement Delangue, Anthony Moi, Pierric Cistac, Tim Rault, Remi Louf, Morgan Funtowicz, Joe Davison, Sam Shleifer, Patrick von Platen, Clara Ma, Yacine Jernite, Julien Plu, Canwen Xu, Teven Le~Scao, Sylvain Gugger, Mariama Drame, Quentin Lhoest, and Alexander Rush.
\newblock Transformers: State-of-the-art natural language processing.
\newblock In Qun Liu and David Schlangen, editors, {\em Proceedings of the 2020 Conference on Empirical Methods in Natural Language Processing: System Demonstrations}, pages 38--45, Online, October 2020. Association for Computational Linguistics.

\bibitem{wu2025agenticreasoningreasoningllms}
Junde Wu, Jiayuan Zhu, and Yuyuan Liu.
\newblock Agentic reasoning: Reasoning llms with tools for the deep research, 2025.

\bibitem{wu2025lessunderstandingchainofthoughtlength}
Yuyang Wu, Yifei Wang, Tianqi Du, Stefanie Jegelka, and Yisen Wang.
\newblock When more is less: Understanding chain-of-thought length in llms, 2025.

\bibitem{wu2025phdknowledgerequiredreasoning}
Zixuan Wu, Francesca Lucchetti, Aleksander Boruch-Gruszecki, Jingmiao Zhao, Carolyn~Jane Anderson, Joydeep Biswas, Federico Cassano, Molly~Q Feldman, and Arjun Guha.
\newblock Phd knowledge not required: A reasoning challenge for large language models, 2025.

\bibitem{xie2024}
Chengxing Xie and Difan Zou.
\newblock A human-like reasoning framework for multi-phases planning task with large language models, 2024.

\bibitem{yang2023intercode}
John Yang, Akshara Prabhakar, Karthik~R Narasimhan, and Shunyu Yao.
\newblock Intercode: Standardizing and benchmarking interactive coding with execution feedback.
\newblock In {\em Thirty-seventh Conference on Neural Information Processing Systems Datasets and Benchmarks Track}, 2023.

\bibitem{yao2023tot}
Shunyu Yao, Dian Yu, Jeffrey Zhao, Izhak Shafran, Thomas~L. Griffiths, Yuan Cao, and Karthik~R Narasimhan.
\newblock Tree of thoughts: Deliberate problem solving with large language models.
\newblock In {\em Thirty-seventh Conference on Neural Information Processing Systems}, 2023.

\bibitem{yao2023react}
Shunyu Yao, Jeffrey Zhao, Dian Yu, Nan Du, Izhak Shafran, Karthik~R Narasimhan, and Yuan Cao.
\newblock React: Synergizing reasoning and acting in language models.
\newblock In {\em The Eleventh International Conference on Learning Representations}, 2023.

\bibitem{ye2025limoreasoning}
Yixin Ye, Zhen Huang, Yang Xiao, Ethan Chern, Shijie Xia, and Pengfei Liu.
\newblock Limo: Less is more for reasoning, 2025.

\bibitem{yu-etal-2018-spider}
Tao Yu, Rui Zhang, Kai Yang, Michihiro Yasunaga, Dongxu Wang, Zifan Li, James Ma, Irene Li, Qingning Yao, Shanelle Roman, Zilin Zhang, and Dragomir Radev.
\newblock {S}pider: A large-scale human-labeled dataset for complex and cross-domain semantic parsing and text-to-{SQL} task.
\newblock In Ellen Riloff, David Chiang, Julia Hockenmaier, and Jun{'}ichi Tsujii, editors, {\em Proceedings of the 2018 Conference on Empirical Methods in Natural Language Processing}, pages 3911--3921, Brussels, Belgium, October-November 2018. Association for Computational Linguistics.

\bibitem{time-scale-o1}
Zhiyuan Zeng, Qinyuan Cheng, Zhangyue Yin, Yunhua Zhou, and Xipeng Qiu.
\newblock Revisiting the test-time scaling of o1-like models: Do they truly possess test-time scaling capabilities?, 2025.

\bibitem{tinyllama}
Peiyuan Zhang, Guangtao Zeng, Tianduo Wang, and Wei Lu.
\newblock Tinyllama: An open-source small language model.
\newblock {\em ArXiv}, abs/2401.02385, 2024.

\bibitem{zhao2025quantifyingcapabilityboundarydeepseek}
Kaikai Zhao, Zhaoxiang Liu, Xuejiao Lei, Jiaojiao Zhao, Zhenhong Long, Zipeng Wang, Ning Wang, Meijuan An, Qingliang Meng, Peijun Yang, Minjie Hua, Chaoyang Ma, Wen Liu, Kai Wang, and Shiguo Lian.
\newblock Quantifying the capability boundary of deepseek models: An application-driven performance analysis, 2025.

\bibitem{ruiqi20}
Ruiqi Zhong, Tao Yu, and Dan Klein.
\newblock Semantic evaluation for text-to-sql with distilled test suite.
\newblock In {\em The 2020 Conference on Empirical Methods in Natural Language Processing}. Association for Computational Linguistics, 2020.

\bibitem{zhou2023language}
Andy Zhou, Kai Yan, Michal Shlapentokh-Rothman, Haohan Wang, and Yu-Xiong Wang.
\newblock Language agent tree search unifies reasoning acting and planning in language models, 2023.

\end{thebibliography}

%\begin{appendices}
%\pagebreak
%\newpage
\appendix
\section{Appendix}
\subsection{Additional Tables}
\begin{table*}[!htbp]
	\small  % or \footnotesize or \scriptsize for smaller font
	\centering
	\renewcommand{\arraystretch}{1.2}
	\setlength{\tabcolsep}{4pt}
	\begin{tabular}{ccccccccccc}
		\toprule
		\multirow{2}{*}{\bf Max New} & \multicolumn{5}{c}{\textbf{Discriminator Performance}} & \multicolumn{5}{c}{\textbf{Reasoning Quality}} \\
		\cmidrule(lr){2-6} \cmidrule(lr){7-11}
		{\bf Token} & \multirow{2}{*}{{PwAcc}} & \multirow{2}{*}{{F1}} & \multirow{2}{*}{{H@1}} & \multirow{2}{*}{{MRR}} & {Fail} & {Rep.} & \multirow{2}{*}{{$1-$TTR}} & {Rep.} & \multirow{2}{*}{{Entropy}} & \multirow{2}{*}{{Diversity}} \\
		& & & & & {Rate} & {Ratio} & & {n-grams} & & \\
		\midrule
		256 & 1.96 & 42.87 & 53.75 & 60.98 & 94.30 & 0.15 & 0.09 & 28.62 & 3.79 & 64.54 \\
		400 & 32.03 & 61.89 & 55.00 & 62.07 & 50.77 & 0.74 & 0.40 & 38.42 & 4.31 & 64.48 \\
		512 & 51.34 & 58.70 & 56.50 & 62.83 & 23.57 & 1.23 & 0.69 & 41.20 & 4.45 & 64.64 \\
		700 & 55.26 & 55.75 & 55.75 & 62.49 & 8.86 & 2.31 & 1.40 & 42.72 & 4.50 & 64.70 \\
		1024 & 56.23 & 55.20 & 55.75 & 62.45 & 6.10 & 3.67 & 2.52 & 43.40 & 4.49 & 64.58 \\
		1536 & 55.99 & 55.28 & 55.50 & 62.32 & 5.92 & 4.60 & 3.57 & 43.78 & 4.47 & 64.42 \\
		2048 & 55.99 & 55.28 & 55.50 & 62.32 & 5.92 & 5.03 & 4.12 & 43.94 & 4.46 & 64.32 \\
		\bottomrule
	\end{tabular}
	%\vspace{-6pt}
	\caption{\label{tab:intrin_eval}
		Performance metrics for different token budgets in intrinsic evaluation with Naive Discrimination setting using Distill-R1.
	}
	%\vspace{-11pt}
\end{table*}

\begin{table*}[!htbp]
	\small
	\centering
	\renewcommand{\arraystretch}{1.2}
	\setlength{\tabcolsep}{4pt}
	\begin{tabular}{lccccccccc}
		\toprule
		\multirow{2}{*}{\bf Models} & \multicolumn{5}{c}{\textbf{Execution Accuracy}} & \multicolumn{1}{c}{\textbf{Exact match}} & \multicolumn{3}{c}{\textbf{Partial match}} \\
		\cmidrule(lr){2-6} \cmidrule(lr){7-7} \cmidrule(lr){8-10}
		& {Easy} & {Mid} & {Hard} & {Extra} & {All} &{Acc} & {Acc} & {Recall} & {F1} \\
		\midrule
		with Logits  & 85.00 & 70.00 & 52.00 & 26.00 & 58.30 & 30.20 & 63.73 & 45.45 & 52.09 \\
		w/o Logits    & 86.00 & 70.00 & 51.00 & 25.00 & 58.00  & 31.80 & 64.83 & 45.34 & 52.25 \\
		\midrule
		Diff (\%)      &  +1.18 & 0.00 & -1.92 & -3.85 & -0.51 & +5.30  & +1.73  & -0.24 & +0.31  \\
		\bottomrule
	\end{tabular}
	%\vspace{-6pt}
	\caption{\label{tab:logits_naive}
		Execution accuracy and gold SQL matching performance in end-to-end evaluation with and w/o logits (+ Executability Check). 
	}
	%\vspace{-11pt}
\end{table*}
\newpage
\begin{table*}[!htbp]
	\small
	\centering
	\renewcommand{\arraystretch}{1.2}
	\setlength{\tabcolsep}{4pt}
	\begin{tabular}{lccccccccc}
		\toprule
		\multirow{2}{*}{\bf Models} & \multicolumn{5}{c}{\textbf{Execution Accuracy}} & \multicolumn{1}{c}{\textbf{Exact match}} & \multicolumn{3}{c}{\textbf{Partial match}} \\
		\cmidrule(lr){2-6} \cmidrule(lr){7-7} \cmidrule(lr){8-10}
		& {Easy} & {Mid} & {Hard} & {Extra} & {All} &{Acc} & {Acc} & {Recall} & {F1} \\
		\midrule
		with Logits  & 85.00 & 71.00 & 50.00 & 18.00 & 56.00 & 29.50 & 65.51 & 43.06 & 50.57 \\
		w/o Logits   & 86.00 & 67.00 & 50.00 & 18.00 & 55.20 & 30.20 & 67.16 & 43.04 & 50.96 \\
		\midrule
		Diff (\%)      & +1.18 & -5.63 & 0.00 & 0.00 & -1.43 & +2.37 & +2.52 & -0.05 & +0.77 \\
		\bottomrule
	\end{tabular}
	%\vspace{-6pt}
	\caption{\label{tab:logits_check}
		Execution accuracy and gold SQL matching performance in end-to-end evaluation with and w/o logits (Naive Discriminator). 
	}
	%\vspace{-11pt}
\end{table*}

\subsection{Example Prompts and Responses}\label{ap1}

\begin{table*}[htbp!]
	%\small
	\scriptsize
	\centering
	\begin{tabular}{p{0.95\linewidth}}
		\toprule
		\texttt{[INST] Answer the following Yes/No question: Is the SQL correct given the utterance?\newline \newline -- Utterance:  What are all distinct countries where singers above age 20 are from?\newline -- SQL:\newline  SELECT DISTINCT country FROM singer WHERE age > 20;\newline -- Answer: [/INST]
		} \\
		\bottomrule
	\end{tabular}
	\caption{\label{tab:example2}
		Example prompt for SQL code evaluation with non-reasoning models.
	}
\end{table*}

 \begin{table*}[htbp!]
 	%\small
 	\scriptsize
 	\centering
 	\begin{tabular}{p{0.95\linewidth}}
 		\toprule
 		\texttt{Determine whether the SQL query is correct or not for the given question. Provide final answer in JSON format with the key 'correct'. For this task do the following:\newline- First, break down the query step by step\newline- Then, determine if the query provides the answer to the given question.\newline You only need to determine in terms of correctness, if query is correct or not. Do not generate alternative query. \newline \newline \#\#\# Question:\newline What are all distinct countries where singers above age 20 are from?\newline \#\#\# SQL query:\newline SELECT DISTINCT country FROM singer WHERE age > 20;\newline \newline <think>\newline Okay, \textcolor{blue}{so I need to figure out if this SQL query is correct for the given question. The question is asking for all distinct countries where singers are above age 20. The SQL query provided is:\newline \newline SELECT DISTINCT country FROM singer WHERE age > 20;\newline \newline Let me break this down step by step.\newline \newline First, the question is about finding countries where singers are above age 20. So, the key here is to find the country (country) where the age (age) is greater than 20.\newline \newline Looking at the SQL query, it starts with SELECT DISTINCT country FROM singer WHERE age > 20. So, it's selecting the country from the singer table where the age is greater than 20.\newline \newline Wait, but the question is about all distinct countries where singers are above age 20. So, the query is correctly selecting the country where the age is greater than 20. It doesn't mention anything about the singers themselves, just the country.\newline \newline So, the query is correct because it's selecting the country where the age is greater than 20, which answers the question about all distinct countries where singers are above age 20.\newline \newline I think that's it. The SQL query is correct.\newline </think>\newline \newline ```json\newline \{\newline  "correct" : true\newline \}\newline '''}
 		} \\
 	\bottomrule
 	\end{tabular}
 	\caption{\label{tab:gen_prompt_text2sql}
 	Example SQL code evaluation with DeepSeek-R1 without using the schema as context. The black text represents the input prompt, while the generated SQL response is highlighted in \textcolor{blue}{blue}.
 	}
\end{table*}

 \begin{table*}[htbp!]
	%\small
	\scriptsize
	%\footnotesize
	\centering
	\begin{tabular}{p{0.95\linewidth}}
		\toprule
		\texttt{
	Determine whether the SQL query is correct or not for the given question. Provide final answer in JSON format with the key 'correct'.\newline  For this task do the following:\newline - First, break down the query step by step. Use the Schema for this.\newline - Then, determine if the query provides the answer to the given question.\newline You only need to determine in terms of correctness, if query is correct or not. Do not generate alternative query.\newline \newline \#\#\# Schema:\newline -- Table stadium: stadium\_id, location, name, capacity, highest, lowest, average\newline -- Table singer: singer\_id, name, country, song\_name, song\_release\_year, age, is\_male\newline -- Table concert: concert\_id, concert\_name, theme, stadium\_id, year\newline -- Table singer\_in\_concert: concert\_id, singer\_id\newline \#\#\# Question:\newline What are all distinct countries where singers above age 20 are from?\newline \#\#\# SQL query:\newline SELECT DISTINCT country FROM singer WHERE age > 20;\newline \newline <think>\newline Okay, \textcolor{blue}{so I need to figure out if the SQL query I wrote is correct for the given question. Let me break it down step by step.\newline \newline First, the question is asking for all distinct countries where singers are above age 20. So, I need to find singers who are older than 20 and then see which countries they come from.\newline \newline Looking at the provided SQL query:\newline SELECT DISTINCT country \newline  FROM singer \newline  WHERE age > 20;\newline \newline Let me analyze each part of this query.\newline \newline 1. **SELECT DISTINCT country**: This part is straightforward. It's asking for the distinct values of the 'country' column. So, it's selecting all unique countries where the condition is met.\newline \newline 2. **FROM singer**: This tells the SQL query where to look for the data. It's selecting from the 'singer' table. So, it's filtering based on the 'singer' table.\newline \newline 3. **WHERE age > 20**: This is the condition that filters the data. It's selecting rows where the 'age' column is greater than 20.\newline \newline Putting it all together, the query is selecting distinct countries from the 'singer' table where the singer's age is greater than 20.\newline \newline Now, let me think about whether this query correctly answers the question.\newline \newline The question is asking for all distinct countries where singers are above age 20. The query does exactly that:\newline \newline - It selects 'country' from 'singer' where 'age' > 20.\newline \newline So, the query is correctly structured to answer the question.\newline \newline I don't see any issues with this query. It's straightforward and meets all the requirements of the question.\newline </think>\newline \newline ```json\newline \{ \newline   "correct": true\newline \} \newline '''}
		} \\
		\bottomrule
	\end{tabular}
	\caption{\label{tab:gen_prompt_text2sql2}
		Example SQL code evaluation with DeepSeek-R1 using the schema as context. The black text represents the input prompt, while the generated SQL response is highlighted in \textcolor{blue}{blue}.
	}
\end{table*}

% SQL: SELECT other_student_details FROM students ORDER BY other_student_details DESC;
 \begin{table*}[htbp!]
	%\small
	\scriptsize
	\centering
	\begin{tabular}{p{0.95\linewidth}}
		\toprule
		\texttt{Answer the question by generating a valid sqlite SQL query. Final answer must start with SELECT.\newline \#\#\# Schema:\newline -- Table addresses: address\_id, line\_1, line\_2, line\_3, city, zip\_postcode, state\_province\_county, country, other\_address\_details\newline -- Table courses: course\_id, course\_name, course\_description, other\_details\newline -- Table departments: department\_id, department\_name, department\_description, other\_details\newline -- Table degree\_programs: degree\_program\_id, department\_id, degree\_summary\_name, degree\_summary\_description, other\_details\newline -- Table sections: section\_id, course\_id, section\_name, section\_description, other\_details\newline -- Table semesters: semester\_id, semester\_name, semester\_description, other\_details\newline -- Table students: student\_id, current\_address\_id, permanent\_address\_id, first\_name, middle\_name, last\_name, cell\_mobile\_number, email\_address, ssn, date\_first\_registered, date\_left, other\_student\_details\newline -- Table student\_enrolment: student\_enrolment\_id, degree\_program\_id, semester\_id, student\_id, other\_details\newline -- Table student\_enrolment\_courses: student\_course\_id, course\_id, student\_enrolment\_id\newline -- Table transcripts: transcript\_id, transcript\_date, other\_details\newline -- Table transcript\_contents: student\_course\_id, transcript\_id\newline \newline \#\#\# Question: What other details can you tell me about students in reverse alphabetical order?\newline \newline <think>\newline Okay, \textcolor{blue}{so I need to figure out what other details I can tell about students in reverse alphabetical order. Let me start by understanding the problem.\newline \newline First, the question is about students ordered alphabetically in reverse, which means from Z to A. I need to find other details about these students. The database schema includes several tables: students, student\_enrolment, student\_enrolment\_courses, etc.\newline \newline I think the key here is to look at the 'students' table because it contains all the necessary information about each student. The 'student\_enrolment' and 'student\_enrolment\_courses' tables might provide additional details, but they seem to be related to how students are enrolled and the courses they take, which might not be directly relevant to their names.\newline \newline So, focusing on the 'students' table, I can get the student\_id, first\_name, last\_name, etc. But the question is about reverse alphabetical order, so I need to sort the students by their names in reverse and then extract the relevant details.\newline \newline I should write a SQL query that selects the necessary fields from the 'students' table, orders them in reverse alphabetical order, and then returns the selected fields. The order by clause should be based on the 'first\_name' column, but I need to make sure it's in reverse.\newline \newline Wait, the 'student\_enrolment' table might have more information about each student, like their course details or other details, but I'm not sure if that's necessary. The 'student\_enrolment\_courses' table seems to be about which courses a student is enrolled in, which might not be directly relevant to their names.\newline \newline So, the main data I can get is from the 'students' table. I can sort them by first\_name in reverse and then select the first\_name, last\_name, etc., to get the details in reverse alphabetical order.\newline \newline I should also consider if there are any other fields that might be useful, like email\_address, ssn, date\_first\_registered, etc., but the question is about other details, so I think focusing on the names is sufficient.\newline \newline Putting it all together, the SQL query should select the first\_name and last\_name from the 'students' table, order them by first\_name in reverse, and then return these fields. This should give me the details in the required order.\newline \newline I think that's the approach. Now, I'll write the SQL query accordingly.\newline </think>\newline \newline To determine the details of students in reverse alphabetical order, we can use the following SQL query:\newline \newline ```sql\newline SELECT first\_name, last\_name\newline FROM students\newline ORDER BY first\_name DESC;\newline ```\newline \newline This query selects the `first\_name` and `last\_name` from the `students` table and orders them in descending alphabetical order. The result will provide the details of students in reverse alphabetical order.}
		} \\
		\bottomrule
	\end{tabular}
	\caption{\label{tab:example3}
		Example SQL code generation with DeepSeek-R1. The black text represents the input prompt, while the generated SQL response is highlighted in \textcolor{blue}{blue}.
	}
\end{table*}

% SQL: SELECT other_student_details FROM students ORDER BY other_student_details DESC;
\begin{table*}[htbp!]
	%\small
	\scriptsize
	\centering
	\begin{tabular}{p{0.95\linewidth}}
		\toprule
		\texttt{[INST] Given database schema and a question in natural language, generate the corresponding SQL query.\newline \newline -- Database student\_transcripts\_tracking:\newline -- Table addresses: address\_id, line\_1, line\_2, line\_3, city, zip\_postcode, state\_province\_county, country, other\_address\_details\newline -- Table courses: course\_id, course\_name, course\_description, other\_details\newline -- Table departments: department\_id, department\_name, department\_description, other\_details\newline -- Table degree\_programs: degree\_program\_id, department\_id, degree\_summary\_name, degree\_summary\_description, other\_details\newline -- Table sections: section\_id, course\_id, section\_name, section\_description, other\_details\newline -- Table semesters: semester\_id, semester\_name, semester\_description, other\_details\newline -- Table students: student\_id, current\_address\_id, permanent\_address\_id, first\_name, middle\_name, last\_name, cell\_mobile\_number, email\_address, ssn, date\_first\_registered, date\_left, other\_student\_details\newline -- Table student\_enrolment: student\_enrolment\_id, degree\_program\_id, semester\_id, student\_id, other\_details\newline -- Table student\_enrolment\_courses: student\_course\_id, course\_id, student\_enrolment\_id\newline -- Table transcripts: transcript\_id, transcript\_date, other\_details\newline -- Table transcript\_contents: student\_course\_id, transcript\_id\newline -- Question: What other details can you tell me about students in reverse alphabetical order?\newline  -- SQL:\newline  [/INST] \textcolor{blue}{ SELECT s.*, a.line\_1, a.line\_2, a.line\_3, a.city, a.zip\_postcode, a.state\_province\_county, a.country, a.other\_address\_details, d.department\_name, dp.degree\_summary\_name, sp.semester\_name\newline  FROM students s\newline  INNER JOIN addresses a ON s.current\_address\_id = a.address\_id\newline  INNER JOIN departments d ON s.department\_id = d.department\_id\newline  INNER JOIN degree\_programs dp ON s.degree\_program\_id = dp.degree\_program\_id\newline  INNER JOIN semesters sp ON s.semester\_id = sp.semester\_id\newline  ORDER BY s.last\_name DESC;\newline \newline [/INST]\newline \newline -- Answer: The students in the student\_transcripts\_tracking database can be sorted in reverse alphabetical order by the last name.\newline		}
		} \\
		\bottomrule
	\end{tabular}
	\caption{\label{tab:example4}
		Example SQL code generation with non-reasoning model (DeepSeek Coder-1.3B). The black text represents the input prompt, while the generated SQL response is highlighted in \textcolor{blue}{blue}.
	}
\end{table*}  

\end{document}